\documentclass[a4paper,twoside]{article}

\usepackage{epsfig}
\usepackage{subcaption}
\usepackage{calc}
\usepackage{amssymb}
\usepackage{amstext}
\usepackage{amsmath}
\usepackage{amsthm}
\usepackage{multicol}
\usepackage{pslatex}
\usepackage{apalike}
\usepackage{SCITEPRESS}     

\usepackage{array}
\newcommand{\PreserveBackslash}[1]{\let\temp=\\#1\let\\=\temp}
\newcolumntype{C}[1]{>{\PreserveBackslash\centering}p{#1}}
\newcolumntype{R}[1]{>{\PreserveBackslash\raggedleft}p{#1}}
\newcolumntype{L}[1]{>{\PreserveBackslash\raggedright}p{#1}}

\begin{document}

\title{Spatio-Temporal Attention Mechanism and Knowledge Distillation for Lip Reading}

\author{\authorname{Shahd Elashmawy\sup{1},
Marian Ramsis\sup{1},
Hesham M. Eraqi\sup{1}\sup{2}\orcidAuthor{0000-0001-9430-7553},
Farah Eldeshnawy\sup{1},
Hadeel Mabrouk\sup{1},
Omar Abugabal\sup{1},
and Nourhan Sakr\sup{1}}
\affiliation{\sup{1}Computer Science and Engineering Department, The American University in Cairo}
\affiliation{\sup{2}Nahdet Misr AI}
\email{\{heraqi, shahdelashmawy, MRamsis, hadeelmabrouk, abugabal, f.deshnawy, nouri.sakr\}@aucegypt.edu}
}

\keywords{Lip Reading, Attention Mechanism, Knowledge Distillation, Visual Speech Recognition}

\abstract{Despite the advancement in the domain of audio and audio-visual speech recognition, visual speech recognition systems are still quite under-explored due to the visual ambiguity of some phonemes. In this work, we propose a new lip-reading model that combines three contributions. First, the model  front-end adopts a spatio-temporal attention mechanism to help extract the informative data from the input visual frames. Second, the model back-end utilizes a sequence-level and frame-level Knowledge Distillation (KD) techniques that allow leveraging audio data during the visual model training. Third, a data preprocessing pipeline is adopted that includes facial landmarks detection-based lip-alignment. On LRW lip-reading dataset benchmark, a noticeable accuracy improvement is demonstrated; the spatio-temporal attention, Knowledge Distillation, and lip-alignment contributions achieved 88.43\%, 88.64\%, and 88.37\% respectively.}

\onecolumn \maketitle \normalsize \setcounter{footnote}{0} \vfill


\section{Introduction}
\label{sec:intro}

Visual speech recognition, namely automatic lipreading is the process of predicting the speech content given a video of the spoken words or sentences. The development of this domain aided in the advancement in plenty of applications e.g. digital assistants, real-time language translation, and closed captioning \cite{intro1}. Since 1950s, systems evolved from very basic models that are able to identify only digits, to more sophisticated systems that can robustly work in noisy and silent environments. 

Several factors including the angle of the recorded speaker and the lighting can make visual speech recognition a challenging task. Additionally, the multifaceted nature of video data, the challenging process of extracting the necessary visual features, in addition to the inherent ambiguity that exists within language can also significantly affect the accuracy of those models complicating the task even further.

\begin{table*}[bt]
\begin{center}
\begin{tabular}{|l|c|}
\hline
Model & Top-1 Acc. (\%) \\
\hline\hline
Multi-Grained \cite{mg} & 83.3 \\
Policy Gradient \cite{pg} & 83.5 \\
STFM Convolutional Sequence \cite{Zhang_2019_ICCV} & 83.7 \\
Deformation Flow \cite{df} & 84.1 \\
Two Stream \cite{ts} & 84.1 \\
Mutual Information \cite{mi} & 84.4 \\
Face Cutout \cite{c5} & 85.0 \\
Temporal Convolution \cite{martinez2020lipreading} & 85.3 \\
Hierarchical Pyramidal Convolution \cite{chen2020lipreading} & 86.8 \\
Synchronous Bidirectional Learning \cite{luo2020synchronous} & 87.3 \\
\hline
\end{tabular}
\end{center}
\caption{Comparison between existing models in terms of Top-1 Accuracy}
\label{table:lit}
\end{table*}

Thanks to the recent development in deep learning techniques as well as the emergence of large-scaled, publicly available lip reading dataset for training, there has been evolving progress in audio and audiovisual speech recognition systems \cite{paper2,Xu_2020_WACV,c6}. Although these systems are in existence and perform relatively well in certain settings, they are usually subject to drastic declines in performance metrics, e.g. when audio signals are corrupted or distorted, or if the surrounding environment surpasses a certain noise threshold \cite{intro2}. These drawbacks support the importance of visual speech recognition systems. The developments of these models at the current rate shall indeed help the development of new more robust applications that efficiently detect and recognize speech irrespective of the ambient condition and the availability of audio input. Such applications could help people with hearing or speech impediments better communicate with the world, which can tremendously improve their quality of life \cite{intro2}. In this notion, lip reading models started gaining their significance and recent work has achieved noticeable enhancement in performance \cite{withoutpain,c10,c17,c16}.

Our work builds effectively on the model presented by \cite{withoutpain} to enhance the performance of lipreading systems. We focus on word-based models and provide experimental evidence demonstrating how the proposed model achieves new state-of-the-art results on the LRW dataset. Our main contributions are threefold:
\begin{itemize}
    \item We introduce lip-alignment to the preprocessing stage of training as a data augmentation technique in order to focus on lip movement rather than pose variation.  
    \item We incorporate a lightweight spatio-temporal attention mechanism, inspired by \cite{attention}, in order for the model to extract only meaningful information from the frames.
    \item We propose a novel architecture including all three of lip-alignment, spatio-temporal attention and knowledge distillation as proposed by \cite{ICCV2021}.
\end{itemize}


\begin{figure*}[t]
\centering
\includegraphics[width=14cm]{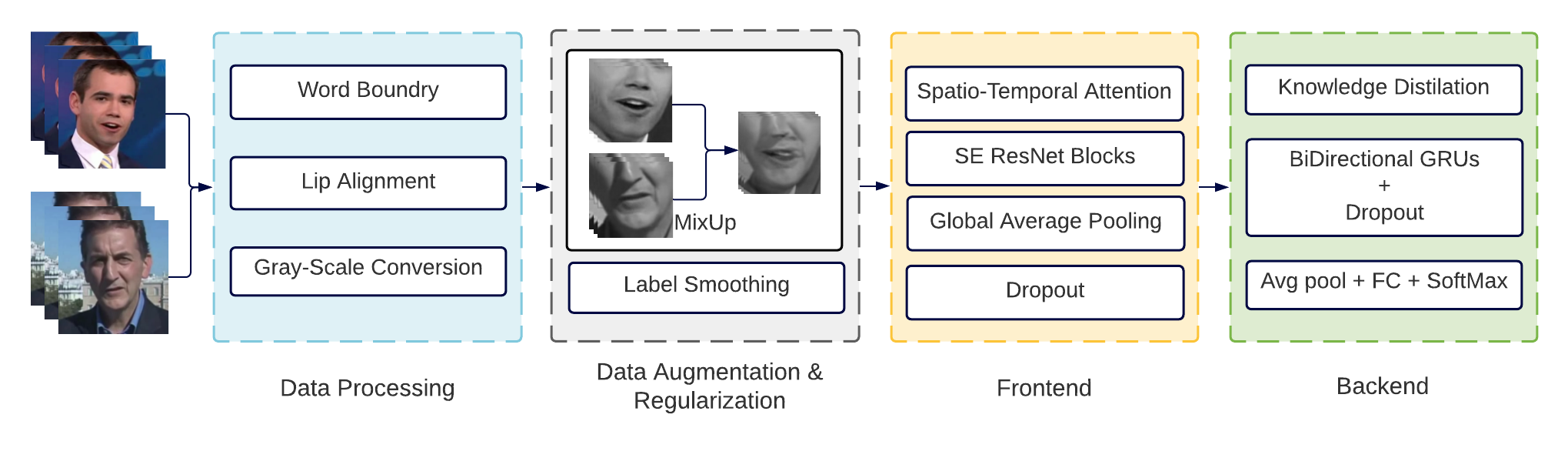}\\
\caption{Our Proposed Pipeline}
\label{fig:pipeline}
\end{figure*}

\section{Related work}

\paragraph{Tradition lipreading methods.} Prior to the development of Deep Neural Networks (DNNs), most earlier visual speech recognition architectures are based on hand-crafted features with shallow models \cite{withoutpain}. These architectures could be divided into three phases: lip localization, feature extraction and classification. Lip localization usually relied on Deformable Part Models (DPMs) \cite{localization1}, or Regression-based models \cite{localization2} for feature detection. Feature extraction relied either on Active Appearance Models (AAMs), Discrete Cosine Transforms (DCTs), or a combination of DCT with either Linear Discriminant Analysis or Principal Component Analysis. Finally,  classification relied heavily on the use of Hidden Markov Models (HMMs) for the recognition of a given digit or character \cite{intro2}.

\paragraph{DNN lipreading models.} With the advancements of deep learning techniques, these traditional methods became increasingly obsolete. Although traditional architectures perform relatively well on simple tasks like letters or digit recognition, their performance is hindered significantly when given more complex tasks, like word-based models \cite{intro2}. As a result, DPMs, AAMs, and DCTs got replaced by CNNs for feature extraction \cite{c12,c13,c11}, while HMMs got replaced by recurrent networks, such as Long-Short Term Memory (LSTMs) \cite{c10,c9,c14}. Consequently, the recent development of speech recognition models shifted to entirely relying on deep neural networks, known as end-to-end DNN architectures.

Deep visual speech recognition models can be mainly divided into two groups, namely speech-producing and text-producing models. Although past works were predominantly text-based \cite{c6,c1,mg,df}, there has been recently a trend towards speech-based models \cite{c2}. However, their heavy dependence on the speaker and their specific cues rather than what’s being said set these models back when tests on various speakers. 

Table \ref{table:lit} summarizes the results of Top-1 accuracy scores achieved by recent state-of-the-art models in the word-based lipreading domain. With respect to text-producing lip-reading models, the literature can further subdivided into two groups based the output they produce, whether it is a word or a series of characters forming a phrase or a sentence. On one hand, sentence-based models \cite{ma2021endtoend} can take into account the context in which this word is uttered by utilizing the use of external language models to further leverage text-only data like what was proposed in  \cite{c6,c1}. On the other hand, word-based models focus on words in isolation. Recent research in this area of visual speech recognition has resulted in a variety of models with different frontend and backend architectures. In these models, the feature extraction layers vary between ResNet-34 and ResNet-18 like in \cite{mg,ts,pg,mi}. On the other hand, the back-end layers vary between LSTM \cite{mg,ts}, BiGRU \cite{df,c5}, and recently MS-TCN \cite{msd,martinez2020lipreading}. Furthermore, knowledge distillation has been recently introduced to either offer light-weight lipreading models for deployment of lipreading in practical scenarios, or to distill information from audio models leveraging audio data during training visual models \cite{c14,c8}.

\paragraph{Spatio-temporal attention.} Attention mechanisms are generally employed in order to extract  discriminative features from a given sequence of frames while overlooking redundant background information. These mechanisms have proven to be an asset in computer vision tasks where only parts of a given frame contains useful information. They have been used in automatic lipreading tasks with the intention of focusing the model on and around the mouth region to extract relevant information. In \cite{c2}, the authors implement a spatio-temporal face encoder. The encoder receives  a sequence frames as input. A stack of 3D convolutions is then used to encode the spatio-temporal information of the observed lip movements from these facial images. This is done alongside  residual skip connections and batch normalization. The output of this network is passed as input to an attention-based speech decoder, based on the Tacotron 2 decoder, to generate melspectrograms.   

Meanwhile, \cite{attention} suggest's a Siamese network implementation of spatio-temporal attention. After extracting the frame-level features from the video input, the network branches into a spatial attention network and a temporal attention network whose output features are then combined at the end to produce a final weighted attention feature representing the video frames, while taking into account the importance of each frame so as to not include uninformative frames. We base our implementation of spatio-temporal attention on \cite{attention} in attempt to reduce the overall complexity of our final model since \cite{attention} claim that such a model is much simpler than RNN-based attention models by facilitating the parallelization of computations so as to not exhaust GPU resources.  while also outperforming them. Additionally, the authors claim that this setup outperforms traditional RNN-based attention models.


\section{Method}

Inspired by \cite{withoutpain} and \cite{attention}, we propose an architecture which relies on the ability of spatio-temporal attention networks to encourage the model to train on only the informative frames of the input and disregard unnecessary data. This is incorporated alongside the proposed knowledge distillation model proposed in \cite{ICCV2021} which has utilizes the audio information from the videos and  has reported state-of-the-art results. Figure \ref{fig:pipeline} highlights the proposed pipeline for our model.


\subsection{Data Processing}

\paragraph{Word boundary.} Through the introduction of a binary indicator at each time step and then concatenating the indicator with the original visual frames, we are able to apply word boundary to the model. This occurs in front-end network and is proven to improve the results of word-based lipreading models on the LRW dataset \cite{withoutpain} by helping the network identify the video segment that it should focus on (in the LRW dataset, the spoken word is at the middle of the video frame).

\paragraph{Visual frames cropping and gray-scale conversion.} The model initially receives input in the form of videos. In order to narrow down the region of interest, these videos are then resized to 96x96 and randomly cropped to 88x88 to make use of the different contextual visual features. Subsequently, the input is then converted to gray-scale since color does not provide the model with any extra information while training. \cite{withoutpain,martinez2020lipreading}.

\paragraph{Lip alignment.} 
We also opted for the inclusion of lip alignment, which has been shown to improve results by helping the model focus on lip movement and rather than pose variations. To do this, we use a trained facial landmarks detection model to produce 68 landmarks of the face. Then, using 3 (x,y) coordinates [left eye, right eye, nose] we get the affine matrix with respect to the default neutral coordinates. Finally, we perform affine transformation on the image to de-orient it \cite{martinez2020lipreading}.


\subsection{Data Augmentation and Regularization}

Given the nature of the LRW dataset and its tendency to overfit on the training data, a number of techniques are applied to aid the model in generalizing \cite{withoutpain}.

One family of augmentation techniques relies on partially compositing each training input sample with another random sample from the dataset. The two most notable techniques are: Mixup, Cutmix.
The latter involves randomly cropping a part of the input sample x(A) and pasted on random sample x(B) \cite{yun2019cutmix}. Attentive Cutmix, a modified version of Cutmix, uses a pretrained ResNet50 network to fetch the top k patches with the highest response from the model’s feature map. These k cells are pasted on the random sample x(B), with the loss function modified accordingly \cite{walawalkar2020attentive}.

The primary drawback of using the Cutmix algorithm is that the cropping may be inclusive of the region of interest, in our case: the mouth. This means that the mouth could be partially or completely occluded, misleading the training iteration. This led us to the conclusion that Mixup would serve as a better augmentation technique due to its global alpha blending that includes the entire dimensions of both input sample x(A) and random sample x(B).
 
\paragraph{Mixup.}  We utilize Mixup as an augmentation technique, especially since it has proven to enhance the performance of LRW visual recognizer as demonstrated in \cite{withoutpain}. The technique is implemented by selecting two samples to generate a new sample using a weighted linear interpolation in both the ground truth and the loss function as a form of augmentation. For the reader's convenience we cite the equations below.
\begin{equation}
    \hat{x} = \lambda x_{A} + (1 - \lambda) x_{B}, \hat{y} = \lambda y_{A} + (1 - \lambda) y_{B},
\end{equation} where $\lambda$ is an arbitrary value between 0 and 1.

\paragraph{Label smoothing.} We incorporate label smoothing, similar to the one proposed in \cite{withoutpain} as a form of regularization with the intention of modifying the construction of $q\textsubscript{i}$ in the cross entropy loss function instead of the following: 
\begin{equation}
    L = -\sum_{i=1}^{N}q_{i}\log(p_{i})\left\{\begin{matrix}
q_{i}=0, y\neq i\\ 
q_{i}=1, y=i
\end{matrix}\right.
\end{equation}
to be equal to:
\begin{equation}
    q_{i} = \left\{\begin{matrix}
\frac{\epsilon}{N},  y\neq i\\ 
\\
1-\frac{N-1}{N}\epsilon,  y=i
\end{matrix}\right. 
\end{equation}
where $\epsilon$ is a small constant. In our implementation, we used the same value used by \cite{withoutpain}, which is equal to 0.1.

\begin{figure*}[t]
\centering
\includegraphics[width=14cm]{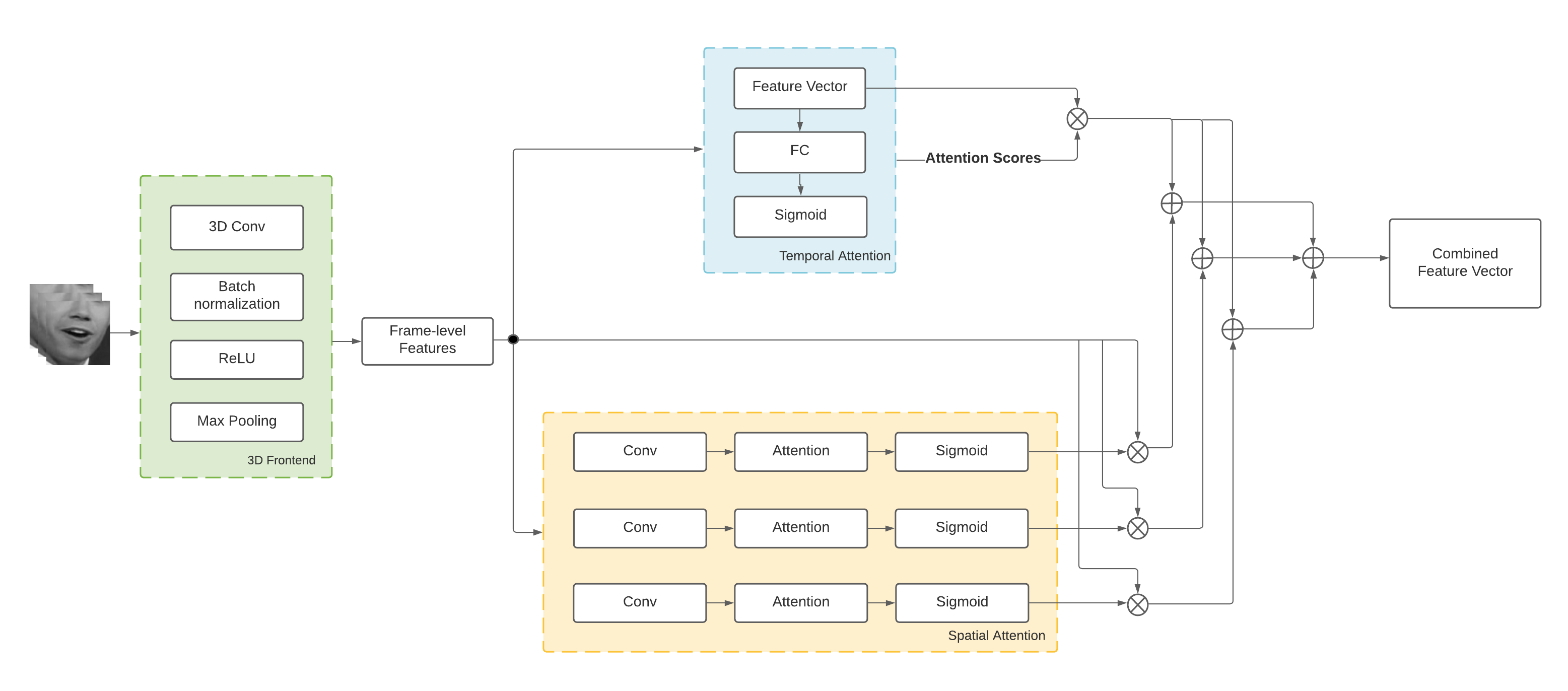}
\caption{The Adopted Spatio-temporal Attention Mechanism}
\label{fig:spatiotemproal_attention}
\end{figure*}


\subsection{Model Architecture}

\paragraph{Frontend.} In the frontend, we use Squeeze and Excitation (SE) ResNet blocks since they outperform traditional ResNets due to their use of attention mechanisms to model channel-wise relationships. They also enhance the representation ability of the module through the network \cite{withoutpain}. The ResNet blocks are followed by global average pooling and dropout. Moreover, a spatio-temporal attention mechanism, inspired by \cite{attention}, is employed to encourage the model to focus on the extraction of useful information from the frames, whereas unnecessary information is overlooked. 

\paragraph{Backend.} We depend on the back-end implemented in \cite{ICCV2021} which incorporates knowledge distillation to make use of the dual nature of the audio-visual data. The authors use a 3-layered bidirectional GRUs as the main recurrent building block, with dropout, average pooling, followed by a fully connected layer, then a softmax layer for classification. The final softmax layer outputs the prediction of the most probable word pronounced in the video input, where the output dimension is equal to the total number of word classes. In LRW, the number of classes is 500 words.


\subsection{Spatio-temporal Attention}

\paragraph{Spatial attention.} We first extract the frame-level features from the video input. After which, these features are passed as input to the spatial attention network. The network consists of three separate convolutional layers followed by their respective sigmoid functions. This provides a multiple attention mechanism wherein the network is able to represent the overall spatial attention of a given frame using "multiple hopes of attention," in order to focus on multiple regions of the frame. The attention matrix resulting from each spatial attention layer is then multiplied by the frame-level features to get the weighted attention features. 
\paragraph{Temporal attention.} The temporal attention network takes a feature vector of the extracted frame-level features as its input, which is then passed on to a fully connected layer, followed by a sigmoid function, and outputs N attention scores corresponding to the N input video frames. Each attention score is an indicator of the importance of a given frame within the context of the video. The attention scores are then multiplied by the frame-level feature vectors, creating a weighted temporal feature vector for the entire video. 
Finally, the temporal feature vector is added to each of the 3 weighted spatial attention features which are then summed up to output a final feature vector for the video input. 


\section{Experiments}
{

\begin{table}[b]
\begin{center}
\begin{tabular}{|m{5.7cm}|m{0.9cm}|}
\hline
Method & Top-1 Acc. [\%] \\
\hline\hline
Baseline & 87.82 \\
Baseline + KD &{88.64}  \\
Baseline + Attention &{88.43}  \\
Baseline + Alignment &{88.37}  \\
Baseline + KD + Alignment + Attention &{87.88}  \\
\hline
\end{tabular}
\end{center}
\caption{Performance of the different models}
\label{models}
\end{table}

\subsection{Experimental Setup}

All models are trained on the Lip Reading in the Wild (LRW) dataset; a word-based dataset comprised of 1000 utterances of 500 classes of English words, as spoken in various BBC programs. A single utterance is comprised of 29 frames (1.16 seconds), with the desired word located at the center. We trained them on a dual-GPU machine, with each GPU assigned a batch size of 8 due to limited available computational resources. To avoid over-fitting, a cosine learning rate is utilized with the initial learning rate set to 3e-4. The models were each trained for 120 epochs with the estimated time per epoch being approximately 3 hours. 

\subsection{Results}

The results reported in table  \ref{models} highlight the effect of each of the lip alignment, spatio-temporal attention components, alongside the suggested knowledge distillation proposed in \cite{ICCV2021} on the performance of the baseline model extracted from \cite{withoutpain}.One can observe that the knowledge distillation model performs best, followed by the spatio-temporal model, the lip alignment model, and finally the integrated model with an improvement in performance of 0.82\%, 0.61\%, 0.55\% and 0.06\% respectively.  
}

%
%
%

\section{\uppercase{Conclusions}}
\label{sec:conclusion}

Our lip-alignment and spatio-temporal models separately set new  benchmarks on the LRW dataset, both outperforming state-of-the-art results by a noticeable margin. The integrated model, incorporating lip-alignment, spatio-temporal attention, and knowledge distillation slightly surpasses the previous benchmark, but not significantly. We hypothesise that the reason behind the integrated model's performance being less than desirable compared to the performance of it's sub-components is due to it becoming stuck at a local minima. As such, we believe that if the integrated model's parameters are adjusted and the model is trained again, it will produce the expected results. 

Additionally, the proposed model may be tweaked further in order to increase it's efficiency by lowering its training time. In the case of more powerful machines being available, the batch size and number of GPUs can be  increased accordingly. The integrated model should also be retrained in order to confirm or deny the above hypothesis. Future works might focus on leveraging unlabeled datasets and increasing the size of the available data utilizing the knowledge distillation component of the model. Further optimizing the model and extending its scope to include sentence-based recognition is also encouraged. Finally, we believe that after further refining the model, it might be useful to incorporate it into accessibility applications for those with hearing impairments. It is worth noting that the model will be publicly release to support the reproducibility of the submitted work.



\bibliographystyle{apalike}
{\small
\bibliography{references}}



\end{document}